\documentclass[preprint]{vgtc}                          
\usepackage{amsmath}
\usepackage{bm}

\graphicspath{{figures/}{pictures/}{images/}{./}} 
\usepackage{times}                     
\usepackage{tabu}                      
\usepackage{booktabs}                  
\usepackage{lipsum}                    
\usepackage{mwe}                       

\usepackage{pifont}
\usepackage{fontawesome5} 
\usepackage{mathptmx}                  
\usepackage{booktabs}     
\usepackage{multirow}     
\usepackage{caption}      
\usepackage{fdsymbol}
\onlineid{1327}
\vgtccategory{Application}

\vgtcinsertpkg
\usepackage{enumitem}
\usepackage{multirow}

\usepackage{float}     

\usepackage[normalem]{ulem}

\title{The Perils of Chart Deception: How Misleading Visualizations Affect Vision-Language Models}

\author{
Ridwan Mahbub\thanks{e-mail: \{rmahbub, saidulis, tahmid20, mizanurr, enamulh\}@yorku.ca} \\
\scriptsize York University
\and
Mohammed Saidul Islam{\footnotemark[1]} \\
\scriptsize York University
\and
Md Tahmid Rahman Laskar{\footnotemark[1]} \\
\scriptsize York University
\and
Mizanur Rahman{\footnotemark[1]} \\
\scriptsize York University
\and
Mir Tafseer Nayeem\thanks{e-mail: mnayeem@ualberta.ca} \\
\scriptsize University of Alberta
\and
Enamul Hoque{\footnotemark[1]} \\
\scriptsize York University
}

\teaser{
  \centering
  \includegraphics[scale = 0.435]{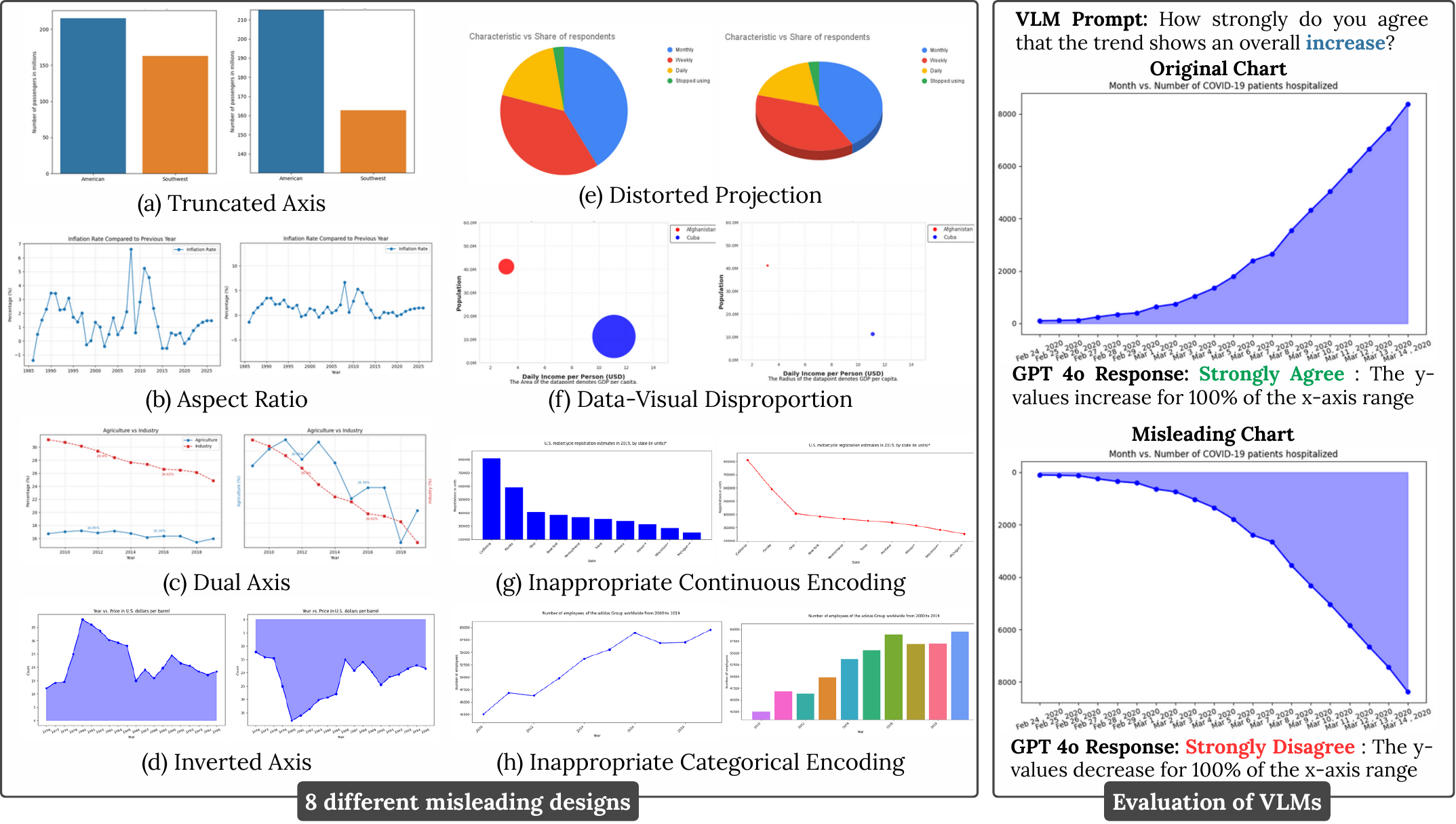}

  \caption{%
  	Overview of the eight misleading chart designs studied: (a) Truncated Axis, (b) Aspect Ratio Distortion, (c) Dual Axis, (d) Inverted Axis, (e) Distorted Projection, (f) Data-Visual Disproportion, (g) Inappropriate Continuous Encoding, and (h) Inappropriate Categorical Encoding. Each pair shows the original (left) and misleading (right) chart. The rightmost section displays GPT-4o outputs for both; in the misleading case, the inverted y-axis leads GPT-4o to wrongly infer a decreasing trend.
  }
  \label{fig:teaser}
  \vspace{-1.5mm}
}

\abstract{%

Information visualizations are powerful tools that help users quickly identify patterns, trends, and outliers, facilitating informed decision-making.  However, when visualizations incorporate deceptive design elements—such as truncated or inverted axes, unjustified 3D effects, or violations of best practices—they can mislead viewers and distort understanding, spreading misinformation. While some deceptive tactics are obvious, others subtly manipulate perception while maintaining a façade of legitimacy. As Vision-Language Models (VLMs) are increasingly used to interpret visualizations, especially by non-expert users, it is critical to understand how susceptible these models are to deceptive visual designs. In this study, we conduct an in-depth evaluation of VLMs' ability to interpret misleading visualizations. By analyzing over 16,000  responses from ten different models across eight distinct types of misleading chart designs, we demonstrate that most VLMs are deceived by them. This leads to altered interpretations of charts, despite the underlying data remaining the same. Our findings highlight the need for robust safeguards in VLMs against visual misinformation.

}


\begin{document}


\firstsection{Introduction}

\maketitle

\begin{figure*}[t!]
    \centering
    \includegraphics[width = \textwidth]{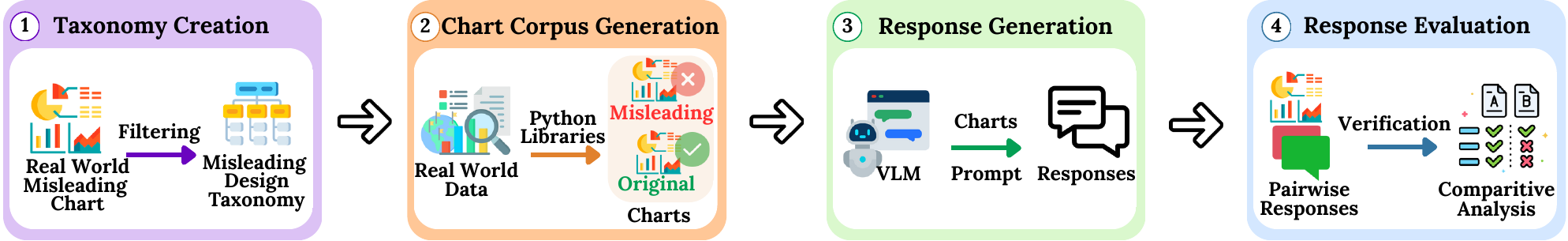} 
\caption{
Overview of our methodology: \textbf{(1)} Taxonomy development, \textbf{(2)} Misleading chart corpus construction, \textbf{(3)} Prompt creation and VLM response generation, \textbf{(4)} Evaluating VLMs by comparing responses across original and misleading charts.
}    
     \label{fig:methodology}
\end{figure*}

Visualizations are powerful tools for transforming complex data into accessible narratives, helping diverse audiences uncover patterns, trends, and anomalies. Across domains, from journalism and public policy to healthcare, finance, and social media, visualizations drive data storytelling and inform high-stakes decisions \cite{segel2010narrative}. However, this communicative power can be a double-edged sword. Subtle manipulations such as truncated axes, skewed aspect ratios, and gratuitous 3D embellishments can produce misleading visualizations that distort perception without altering the underlying data \cite{huff2023lie, cairo2019charts, 10.1145/3613904.3642448}. These practices don’t fabricate facts; instead, they subtly alter their visual representation to amplify or minimize perceived differences, influencing narratives and decisions \cite{huff2023lie, cairo2019charts, 10.1145/3613904.3642448}. Figure \ref{fig:teaser} illustrates \emph{eight} such design tactics studied in this work. For instance, in Figure \ref{fig:teaser}(b), an inflation trend is made to look significantly flatter by changing the aspect ratio—an alteration that may go unnoticed but can fundamentally mislead the viewer. Such distortions are dangerous when they influence public discourse or policy understanding without the audience being aware \cite{lauer2020deceptive}. 

The risks are even greater when these visualizations are used in real-world systems. In finance, healthcare, policymaking, and enterprise reporting, visualizations interpreted by AI systems help automate investment decisions, risk assessment, strategic planning, and public communication \cite{stadler2016improving, uddin2024data}. On social media, they influence public discourse \cite{chen2017social}. If these systems rely on  Vision-Language Models (VLMs) that are susceptible to misleading visuals, the consequences can be serious—leading to poor financial decisions, regulatory mistakes, or the spread of misinformation. 

As VLMs become increasingly embedded in data analysis workflows, their ability to interpret visualizations faithfully is both essential and urgent. VLMs have shown remarkable promise in tasks such as chart summarization \cite{kantharaj2022chart}, question answering \cite{hoque2022chart}, visualization generation \cite{han2023chartllama}, and improving chart accessibility for individuals with visual impairments or low visual literacy \cite{choe2024enhancing, gorniak2024vizability}. Given their growing role in analysis and accessibility, it is essential that VLMs can accurately interpret data, even when visualizations are designed to deceive. Previous research has explored VLMs’ capacity to characterize misleading charts, but often under idealized conditions where prompts explicitly hint at the deception \cite{alexander2024can, lo2024good}. In everyday use, however, people usually engage with visualizations through neutral prompts, unaware of potential misrepresentation \cite{hoque2022chart}. This gap between experimental conditions and real-world use leads to a key question: Can VLMs accurately detect and resist misleading charts in the absence of any explicit signals?

In this study, we examine this overlooked problem: \emph{How do deceptive but statistically valid chart designs impact VLM reasoning when no cues about deception are provided?}  We systematically test whether VLMs fall for common visual tricks under realistic, neutral prompting conditions—focusing on two questions:\\
\noindent \textbf{RQ1:} Do subtle misleading chart designs influence the responses of VLMs, and if so, are certain types of visual deception more effective at misleading these models than others?\\ 
\noindent \textbf{RQ2:} Are all VLMs equally susceptible to visual deception, or do some models demonstrate greater robustness? 

To answer these, we present the first large-scale evaluation of VLM robustness to misleading chart designs, making the following contributions: (\textbf{1}) We introduce a taxonomy of eight widely used deceptive chart tactics and build a benchmark of 1,600 charts—originals and misleading variants—spanning diverse visual distortions. (\textbf{2}) We evaluate 10 cutting-edge VLMs (proprietary and open-source), generating over 16,000 responses to assess model susceptibility. (\textbf{3}) Our results reveal that even top models are vulnerable to subtle visual manipulations, highlighting serious risks for real-world deployment.Finally, 
our code and dataset are made publicly available at \href{https://github.com/vis-nlp/visDeception}{https://github.com/vis-nlp/visDeception}.

\section{Related Work}
\subsection{Misleading Visual Designs}

The potential for charts to mislead has been recognized since the 1950s, with classic works like \emph{``How to Lie with Statistics''}~\cite{huff2023lie} illustrating how visual and statistical representations can distort information and contribute to misinformation. Researchers have examined this issue from multiple perspectives. Some studies have investigated deception strategies in common chart types, i.e., graphs, lines, and pie, 
through user studies assessing their impact on interpretation and decision-making~\cite{pandey2015deceptive, lauer2020deceptive}. Lan \textit{et al.} conducted five focus group studies to investigate the underlying reasons behind visualization design flaws \cite{lan2024came}. Ge \textit{et al.} proposed a definition of visualization literacy that includes the ability to recognize visual deception, arguing that this skill is a vital component of overall visual literacy \cite{ge2023calvi}.  Lisnic \textit{et al.} \cite{lisnic2023misleading, 10.1145/3613904.3642448} analyzed various tactics used to spread misleading narrative using deceptive visualization in social media. Some studies have focused on very specific aspects of visual deception tactics, such as axis truncation \cite{correll2020truncating, long2024cut}. On the other hand, Fan \textit{et al.} \cite{fan2022annotating} explored techniques to mitigate such misleading designs in line charts. While the effects of misleading charts on human interpretation have been widely studied, their impact on VLMs remains largely unexplored.

\subsection{Language Models for Chart Understanding}

Significant improvements have been made in chart reasoning and understanding due to the advances in large vision-language models \cite{islam-etal-2024-datanarrative}. These models can be categorized into: general-purpose multimodal models and chart-specific models. The general-purpose multimodal models can be closed-source or open-source. The closed-source models~\cite{achiam2023gpt, geminiteam2024gemini15unlockingmultimodal} achieve the state-of-the-art performance on chart understanding benchmarks ~\cite{masry2022chartqa, wang2024charxiv} and often outperform their open-source counterparts ~\cite{qwen2.5, li2024llava, chen2024expanding, wu2024deepseek, abdin2024phi}. However, the open-source ones are also making rapid progress to close the gap. For instance, the possibility of fine-tuning open-source models has led to the development of various chart-specific models ~\cite{masry-etal-2025-chartgemma, masry-etal-2024-chartinstruct, zhang2024tinychart,masry2023unichartuniversalvisionlanguagepretrained}, which demonstrate strong performance on various benchmarks ~\cite{masry2022chartqa, akhtar2023chartcheck, kantharaj2022chart}. While existing VLMs perform very well on chart understanding tasks like ChartQA \cite{masry2022chartqa}; their ability to detect and accurately respond to questions about various types of misleading charts remains largely under-explored.

For interpreting misleading charts, while recent studies have examined the ability of language models, the evaluation contexts often diverge from the conditions under which such visuals are encountered in real-world scenarios. For example, Lo et al. \cite{lo2024good} was the first to investigate the detection of misleading charts using real-world examples; however, the prompts provided to the models often implicitly signaled the presence of misleading elements. Alexander et al. \cite{alexander2024can} conducted a similar study on a different corpus consisting of misleading tweets. They use both the tweet text and the attached chart with the tweet to determine if the tweet is misleading or not. These setups contrast with real-world scenarios, where users typically remain unaware that a chart may be deceptive and simply pose questions about the chart without any suspicion of misinformation. Our study addresses the effects of misleading designs on VLMs in this more naturalistic setting.

\section{Methodology}
Our methodology follows four stages to investigate whether and how misleading chart designs affect the performance of VLMs (Figure~\ref{fig:methodology}). First, we develop a taxonomy of common misleading chart designs. Second, we construct a chart corpus containing both controlled (accurate) and misleading versions of each chart. Third, we design prompts to systematically examine how different misleading designs impact model responses (\textbf{RQ1}). Finally, we develop an evaluation framework to identify which VLMs are more vulnerable to these misleading designs  (\textbf{RQ2}). Below, we detail each stage.

\subsection{Taxonomy Creation}
We began by reviewing prior work on real-world misleading visualizations, which often involve design-based deception alongside misleading captions or framing \cite{lo2022misinformed, lisnic2023misleading}. Inspired by these examples, we identified recurring patterns of chart designs that, while statistically accurate, can still distort interpretation, such as inverting an axis to reverse the perceived trend. To ensure systematic evaluation, we refined our focus to include only those misleading designs compatible with survey-style, generalized questions. This choice enabled consistent and controlled testing of VLM susceptibility, following the methodology of Lauer \textit{et al.} \cite{lauer2020deceptive}. The selected chart types also allowed us to pair each visualization with generic, reusable questions—for instance, in a bar chart, a simple prompt comparing the heights of two bars can reliably reveal the effect of a misleading design, without the need for chart-specific question engineering. To maintain this level of generality and control, we excluded forms of multimodal deception (e.g., misleading captions or annotations) and complex multi-view dashboards, where consistent survey-style prompts would be difficult to apply.

The final taxonomy is shown in Table~\ref{tab:taxonomy}. Initially, we categorize misleading chart designs into two broad categories: \textit{(a)} \textbf{Axis Manipulation}—alterations to chart axes that distort data implications, and
\textit{(b)} \textbf{Misleading Encoding}—visual encodings that deviate from best practices and misrepresent trends or magnitudes. Each category includes four representative types of visual distortions that can potentially mislead users. These designs have been widely recognized for their deceptive potential in prior studies \cite{lauer2020deceptive, pandey2015deceptive, szafir2018good}.

\begin{table}[t]
\centering
\renewcommand{\arraystretch}{1.3} 
\caption{Taxonomy of misleading chart designs we evaluated.}
\label{tab:taxonomy}
\resizebox{8.2cm}{!}{%
\begin{tabular}{lll}
\toprule[1pt]
\textbf{Category} & \textbf{Misleading Design} & \textbf{Chart Type} \\ \hline
 & \cellcolor[HTML]{EFEFEF}\texttt{Truncated Axis} \ [Fig. \ref{fig:teaser} (a)] & \cellcolor[HTML]{EFEFEF}Bar \\ \cline{2-3} 
 & \texttt{Aspect Ratio} \ [Fig. \ref{fig:teaser} (b)] & Line \\ \cline{2-3} 
 & \cellcolor[HTML]{EFEFEF}\texttt{Dual Axis} \ [Fig. \ref{fig:teaser} (c)] & \cellcolor[HTML]{EFEFEF}Line \\ \cline{2-3} 
\multirow{-4}{*}{Axis Manipulation} &\texttt{Inverted Axis} \ [Fig. \ref{fig:teaser} (d)] & Line \\ \hline
 & \cellcolor[HTML]{EFEFEF}\texttt{Distorted Projection}  \ [Fig. \ref{fig:teaser} (e)] & \cellcolor[HTML]{EFEFEF}Pie \\ \cline{2-3} 
 &\texttt{Data Visual Disproportion}   \ [Fig. \ref{fig:teaser} (f)] & Bubble \\ \cline{2-3} 
 & \cellcolor[HTML]{EFEFEF}\texttt{Inappropriate Continuous Encoding}    \ [Fig. \ref{fig:teaser} (g)] & \cellcolor[HTML]{EFEFEF}Bar \\ \cline{2-3} 
\multirow{-4}{*}{Misleading Encoding} &\texttt{Inappropriate Categorical Encoding}     \ [Fig. \ref{fig:teaser} (h)] & Line \\ \toprule[1pt]
\end{tabular}%
}

\end{table}

The subcategories of Axis Manipulation include: \textit{(i)} \textbf{Truncated Axis}: \textit{y-axis} does not start at zero, exaggerating differences between bars (Figure \ref{fig:teaser}(a)),
\textit{(ii)} \textbf{Aspect Ratio}: scales are distorted, misleading the viewer about the rate of change (Figure \ref{fig:teaser}(b)), 
\textit{(iii)} \textbf{Dual Axis}: two variables with different scales are plotted against separate \textit{y-axes}, often suggesting a spurious correlation (Figure \ref{fig:teaser}(c)), 
\textit{(iv)} \textbf{Inverted Axis}: \textit{Y-axis} is flipped, which can reverse the perceived trend direction (Figure \ref{fig:teaser}(d)).

The subcategories of misleading encoding include:  
\textit{(i)} \textbf{Distorted Projection}: perspective effects make some segments of a 3D chart appear larger than they are (Figure \ref{fig:teaser}(e)), \textit{(ii)} \textbf{Data-Visual Disproportion}: bubble size is mapped to \textit{radius} instead of \textit{area}, visually inflating differences (Figure \ref{fig:teaser}(f)), \textit{(iii)} \textbf{Inappropriate Continuous Encoding}: a line chart is used to plot categorical variables, falsely implying continuity or trend (Figure \ref{fig:teaser}(g)), \textit{(iv)} \textbf{Inappropriate Categorical Encoding}: continuous data is displayed using bar charts, disrupting perception of gradual variation (Figure \ref{fig:teaser}(h)).

\subsection{Chart Corpus Construction}

While previous studies \cite{lo2022misinformed} have explored misleading charts, a major limitation is the absence of chart benchmarks containing both misleading charts and their corresponding original versions. Such paired examples are critical for isolating the impact of misleading design choices on VLMs. By controlling all other variables and altering only the chart design, we can directly attribute any differences in model responses to the misleading elements. Therefore, we first construct a corpus consisting of 800 original charts (100 charts of each of the 8 misleading subcategories), each paired with a misleading variant using the data tables from the benchmark ChartQA \cite{masry2022chartqa} corpus. We utilize widely adopted Python libraries \cite{Hunter:2007, Waskom2021} to generate both the original and misleading chart images. We ensured that both chart images (original and misleading) maintained consistent visual styles (e.g., chart type, color scheme, layout, typography), isolating the impact of the misleading design alone.

\begin{figure}[t]
    \centering
    \includegraphics[width=\columnwidth]{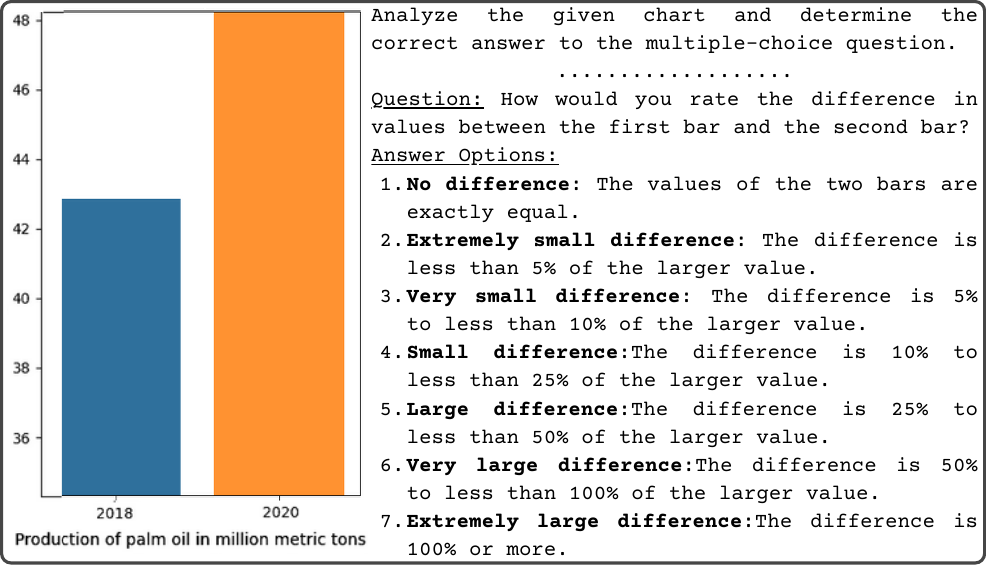} 
    \caption{An example response prompt featuring a bar chart (\textit{truncated axis}), formatting instructions, and precise QA options.}
     \label{fig:prompt}
\end{figure}

\begin{table*}[t]
\centering
\caption{Results of the Wilcoxon signed-rank test comparing original and misleading chart pairs. Here, $p < 0.05$ (in \textbf{bold} font) demonstrates that the model is significantly influenced by the misleading chart design, and the corresponding $z$ value indicates the magnitude of this effect.}
\label{tab:result}
\resizebox{17.5cm}{!} 
{
\begin{tabular}{lcccccccccccccccc}
\toprule[1pt]
\multicolumn{1}{l|}{} & \multicolumn{2}{c}{\textbf{\begin{tabular}[c]{@{}c@{}}Aspect\\ Ratio\end{tabular}}} & \multicolumn{2}{c}{\textbf{\begin{tabular}[c]{@{}c@{}}Distorted\\ Projection\end{tabular}}} & \multicolumn{2}{c}{\textbf{\begin{tabular}[c]{@{}c@{}}Dual\\ Axis\end{tabular}}} & \multicolumn{2}{c}{\textbf{\begin{tabular}[c]{@{}c@{}}Inappropriate \\ Categorical Enc.\end{tabular}}} & \multicolumn{2}{c}{\textbf{\begin{tabular}[c]{@{}c@{}}Inappropriate\\ Continuous Enc.\end{tabular}}} & \multicolumn{2}{c}{\textbf{\begin{tabular}[c]{@{}c@{}}Inverted\\ Axis\end{tabular}}} & \multicolumn{2}{c}{\textbf{\begin{tabular}[c]{@{}c@{}}Data Visual\\ Disproportion\end{tabular}}} & \multicolumn{2}{c}{\textbf{\begin{tabular}[c]{@{}c@{}}Truncated\\ Axis\end{tabular}}} \\ \cline{2-17} 
\multicolumn{1}{l|}{\multirow{-2}{*}{\textbf{Models}}} & z value & \multicolumn{1}{c|}{$p$} & z value & \multicolumn{1}{c|}{$p$} & z value & \multicolumn{1}{c|}{$p$} & z value & \multicolumn{1}{c|}{$p$} & z value & \multicolumn{1}{c|}{$p$} & z value & \multicolumn{1}{c|}{$p$} & z value & \multicolumn{1}{c|}{$p$} & z value & $p$ \\ \toprule[1pt]
\multicolumn{17}{l}{\textit{\faLock \; Closed-Source Models}} \\
\rowcolor[HTML]{EFEFEF} 
\multicolumn{1}{l|}{\cellcolor[HTML]{EFEFEF}GPT-4o} & -3.25 & \multicolumn{1}{c|}{\cellcolor[HTML]{EFEFEF}$\bm{9.6e^{-4}}$} & -0.80 & \multicolumn{1}{c|}{\cellcolor[HTML]{EFEFEF}0.41} & -3.10 & \multicolumn{1}{c|}{\cellcolor[HTML]{EFEFEF}$\bm{1.6e^{-3}}$} & -1.26 & \multicolumn{1}{c|}{\cellcolor[HTML]{EFEFEF}0.19} & -8.14 & \multicolumn{1}{c|}{\cellcolor[HTML]{EFEFEF}$\bm{9.1e^{-17}}$} & -7.78 & \multicolumn{1}{c|}{\cellcolor[HTML]{EFEFEF}$\bm{1.3e^{-15}}$} & -2.27 & \multicolumn{1}{c|}{\cellcolor[HTML]{EFEFEF}$\bm{1.9e^{-2}}$} & -7.61 & $\bm{6.4e^{-15}}$ \\
\rowcolor[HTML]{EFEFEF} 
\multicolumn{1}{l|}{\cellcolor[HTML]{EFEFEF}GPT-4o-mini} & -7.41 & \multicolumn{1}{c|}{\cellcolor[HTML]{EFEFEF}$\bm{7.3e^{-14}}$} & -0.44 & \multicolumn{1}{c|}{\cellcolor[HTML]{EFEFEF}0.63} & -3.03 & \multicolumn{1}{c|}{\cellcolor[HTML]{EFEFEF}$\bm{2.0e^{-3}}$} & -0.29 & \multicolumn{1}{c|}{\cellcolor[HTML]{EFEFEF}0.75} & -6.96 & \multicolumn{1}{c|}{\cellcolor[HTML]{EFEFEF}$\bm{3.8e^{-13}}$} & -7.97 & \multicolumn{1}{c|}{\cellcolor[HTML]{EFEFEF}$\bm{2.8e^{-16}}$} & -2.18 & \multicolumn{1}{c|}{\cellcolor[HTML]{EFEFEF}$\bm{1.7e^{-4}}$} & -6.45 & $\bm{1.3e^{-12}}$ \\
\rowcolor[HTML]{EFEFEF} 
\multicolumn{1}{l|}{\cellcolor[HTML]{EFEFEF}Gemini-2.5-Pro} & -2.97 & \multicolumn{1}{c|}{\cellcolor[HTML]{EFEFEF}$\bm{6.6e^{-4}}$} & -0.83 & \multicolumn{1}{c|}{\cellcolor[HTML]{EFEFEF}0.39} & -1.53 & \multicolumn{1}{c|}{\cellcolor[HTML]{EFEFEF}0.12} & -0.18 & \multicolumn{1}{c|}{\cellcolor[HTML]{EFEFEF}0.55} & -1.5 & \multicolumn{1}{c|}{\cellcolor[HTML]{EFEFEF}0.10} & -2.8 & \multicolumn{1}{c|}{\cellcolor[HTML]{EFEFEF}$\bm{1.9e^{-3}}$} & -0.15 & \multicolumn{1}{c|}{\cellcolor[HTML]{EFEFEF}0.87} & -5.6 & $\bm{6.8e^{-9}}$ \\
\rowcolor[HTML]{EFEFEF} 
\multicolumn{1}{l|}{\cellcolor[HTML]{EFEFEF}Claude-3.7-Sonnet} & -0.59 & \multicolumn{1}{c|}{\cellcolor[HTML]{EFEFEF}0.54} & -2.00 & \multicolumn{1}{c|}{\cellcolor[HTML]{EFEFEF}$\bm{3.3e^{-2}}$} & -4.34 & \multicolumn{1}{c|}{\cellcolor[HTML]{EFEFEF}$\bm{9.3e^{-6}}$} & -1.27 & \multicolumn{1}{c|}{\cellcolor[HTML]{EFEFEF}0.17} & -4.45 & \multicolumn{1}{c|}{\cellcolor[HTML]{EFEFEF}$\bm{6.7e^{-6}}$} & -6.99 & \multicolumn{1}{c|}{\cellcolor[HTML]{EFEFEF}$\bm{1.2e^{-12}}$} & -2.44 & \multicolumn{1}{c|}{\cellcolor[HTML]{EFEFEF}$\bm{1.2e^{-2}}$} & -3.33 & $\bm{2.2e^{-4}}$ \\
\rowcolor[HTML]{EFEFEF} 
\multicolumn{1}{l|}{\cellcolor[HTML]{EFEFEF}Claude-3.5-Haiku} & -6.21 & \multicolumn{1}{c|}{\cellcolor[HTML]{EFEFEF}$\bm{2.3e^{-11}}$} & -1.30 & \multicolumn{1}{c|}{\cellcolor[HTML]{EFEFEF}0.17} & -4.31 & \multicolumn{1}{c|}{\cellcolor[HTML]{EFEFEF}$\bm{3.7e^{-6}}$} & -0.15 & \multicolumn{1}{c|}{\cellcolor[HTML]{EFEFEF}0.87} & -7.32 & \multicolumn{1}{c|}{\cellcolor[HTML]{EFEFEF}$\bm{1.5e^{-13}}$} & -7.58 & \multicolumn{1}{c|}{\cellcolor[HTML]{EFEFEF}$\bm{1.3e^{-14}}$} & -2.00 & \multicolumn{1}{c|}{\cellcolor[HTML]{EFEFEF}$\bm{3.0e^{-2}}$} & -3.54 & $\bm{9.3e^{-5}}$ \\
\midrule
\multicolumn{17}{l}{\textit{\faLockOpen \; Open-Source Models}} \\
\rowcolor[HTML]{DBF7FF} 
\multicolumn{1}{l|}{\cellcolor[HTML]{DBF7FF}LLaVA-v1.6-Mistral-7B} & -2.02 & \multicolumn{1}{c|}{\cellcolor[HTML]{DBF7FF}$\bm{2.5e^{-2}}$} & -2.63 & \multicolumn{1}{c|}{\cellcolor[HTML]{DBF7FF}$\bm{4.9e^{-3}}$} & -1.00 & \multicolumn{1}{c|}{\cellcolor[HTML]{DBF7FF}0.31} & -5.09 & \multicolumn{1}{c|}{\cellcolor[HTML]{DBF7FF}$\bm{2.8e^{-8}}$} & -8.59 & \multicolumn{1}{c|}{\cellcolor[HTML]{DBF7FF}$\bm{4.2e^{-23}}$} & -5.44 & \multicolumn{1}{c|}{\cellcolor[HTML]{DBF7FF}$\bm{3.8e^{-10}}$} & -6.90 & \multicolumn{1}{c|}{\cellcolor[HTML]{DBF7FF}$\bm{3.7e^{-13}}$} & -1.00 & 0.32 \\
\rowcolor[HTML]{DBF7FF} 
\multicolumn{1}{l|}{\cellcolor[HTML]{DBF7FF}InternLM-7B} & -6.27 & \multicolumn{1}{c|}{\cellcolor[HTML]{DBF7FF}$\bm{4.2e^{-11}}$} & -2.02 & \multicolumn{1}{c|}{\cellcolor[HTML]{DBF7FF}$\bm{3.9e^{-2}}$} & -1.57 & \multicolumn{1}{c|}{\cellcolor[HTML]{DBF7FF}0.11} & -0.03 & \multicolumn{1}{c|}{\cellcolor[HTML]{DBF7FF}0.97} & -1.57 & \multicolumn{1}{c|}{\cellcolor[HTML]{DBF7FF}$\bm{9.7e^{-2}}$} & -6.45 & \multicolumn{1}{c|}{\cellcolor[HTML]{DBF7FF}$\bm{1.2e^{-13}}$} & -1.00 & \multicolumn{1}{c|}{\cellcolor[HTML]{DBF7FF}0.32} & -1.00 & 0.32 \\
\rowcolor[HTML]{DBF7FF} 
\multicolumn{1}{l|}{\cellcolor[HTML]{DBF7FF}Phi-4-5.57B} & -4.33 & \multicolumn{1}{c|}{\cellcolor[HTML]{DBF7FF}$\bm{1.3e^{-5}}$} & -1.34 & \multicolumn{1}{c|}{\cellcolor[HTML]{DBF7FF}0.15} & -0.73 & \multicolumn{1}{c|}{\cellcolor[HTML]{DBF7FF}0.46} & -4.07 & \multicolumn{1}{c|}{\cellcolor[HTML]{DBF7FF}$\bm{2.8e^{-5}}$} & -2.06 & \multicolumn{1}{c|}{\cellcolor[HTML]{DBF7FF}$\bm{3.4e^{-2}}$} & -7.18 & \multicolumn{1}{c|}{\cellcolor[HTML]{DBF7FF}$\bm{1.8e^{-15}}$} & -4.38 & \multicolumn{1}{c|}{\cellcolor[HTML]{DBF7FF}$\bm{5.3e^{-7}}$} & -5.5 & $\bm{3.9e^{-10}}$ \\
\rowcolor[HTML]{DBF7FF} 
\multicolumn{1}{l|}{\cellcolor[HTML]{DBF7FF}Qwen2.5-VL-7B} & -6.95 & \multicolumn{1}{c|}{\cellcolor[HTML]{DBF7FF}$\bm{2.43e^{-13}}$} & -1.33 & \multicolumn{1}{c|}{\cellcolor[HTML]{DBF7FF}0.18} & -3.48 & \multicolumn{1}{c|}{\cellcolor[HTML]{DBF7FF}$\bm{1.9e^{-4}}$} & -5.05 & \multicolumn{1}{c|}{\cellcolor[HTML]{DBF7FF}$\bm{3.8e^{-7}}$} & -2.52 & \multicolumn{1}{c|}{\cellcolor[HTML]{DBF7FF}$\bm{1.03e^{-52}}$} & -7.29 & \multicolumn{1}{c|}{\cellcolor[HTML]{DBF7FF}$\bm{2.8e^{-15}}$} & -1.22 & \multicolumn{1}{c|}{\cellcolor[HTML]{DBF7FF}0.16} & -5.96 & $\bm{1.8e^{-9}}$ \\
\rowcolor[HTML]{DBF7FF} 
\multicolumn{1}{l|}{\cellcolor[HTML]{DBF7FF}MiniCPM-V2.6-8B} & -4.56 & \multicolumn{1}{c|}{\cellcolor[HTML]{DBF7FF}$\bm{4.3e^{-6}}$} & -0.14 & \multicolumn{1}{c|}{\cellcolor[HTML]{DBF7FF}0.89} & -0.86 & \multicolumn{1}{c|}{\cellcolor[HTML]{DBF7FF}0.38} & -1.90 & \multicolumn{1}{c|}{\cellcolor[HTML]{DBF7FF}$5.5e^{-2}$} & -3.83 & \multicolumn{1}{c|}{\cellcolor[HTML]{DBF7FF}$\bm{1.02e^{-4}}$} & -7.35 & \multicolumn{1}{c|}{\cellcolor[HTML]{DBF7FF}$\bm{1.3e^{-14}}$} & -0.69 & \multicolumn{1}{c|}{\cellcolor[HTML]{DBF7FF}0.44} & -1.91 & $5.2e^{-2}$ \\
\toprule[1pt]
\end{tabular}
}
\end{table*}

\subsection{Response Generation}
The response generation phase consists of two main components: \textit{(i) Prompt Construction} and \textit{(ii) Model Selection}.

\textbf{(i) Prompt Construction:}  
To evaluate VLM responses across various misleading chart types, we design a prompt $P$ comprising three elements: \textit{(a)} a general instruction $f$, \textit{(b)} a chart-specific question $C$, and \textit{(c)} a chart image $I$ (original or misleading). The instruction $f$ defines the task scope and explicit formatting guidelines for the response. The question $C$ is a multiple-choice item about the chart, requiring the model to select an answer on a 7-point Likert scale~\cite{likert1932technique} alongside a justification for its choice. To capture differences across misleading designs, we define a distinct chart-specific question $c_{\text{type}} \in C$ for each misleading chart type.

Figure~\ref{fig:prompt} illustrates an example prompt used in our study, which includes a general instruction, a misleading bar chart exhibiting a \textit{truncated axis}, and a multiple-choice question. The instructions direct the model to select the most appropriate answer from the listed options and justify its selection in a structured manner. Our prompt design aims to assess whether VLMs are susceptible to misleading visualizations. Ideally, a robust model should recognize the misleading element in the chart, provide consistent answers across both the original and misleading versions, and articulate a clear rationale explaining why the chart is misleading, if applicable.

Each question in our benchmark is designed to reveal the specific effect of the corresponding misleading chart design. We adapt and extend the question formulations from Lauer et al.~\cite{lauer2020deceptive} by incorporating a neutral response option to allow nuanced interpretations. The following guidelines inform our question generation strategy:

\noindent $\bullet$  \textbf{Truncated axis (bar charts):} As this design amplifies visual differences of values, questions focus on comparing bar magnitudes.\\
\noindent$\bullet$  \textbf{Trend distortions (line charts; e.g., aspect ratio, inverted axes):} Questions evaluate perceived trend direction and strength.\\
\noindent$\bullet$  \textbf{Distorted projections and visual disproportion (e.g., pie charts, area charts):} Questions prompt models to compare relative magnitudes of chart segments.\\
\noindent$\bullet$  \textbf{Inappropriate data encoding (e.g., continuous vs. categorical mismatches):} Questions examine whether the model erroneously infers trends in discretized continuous data or fails to recognize categorical data where no trend should be inferred.
    
\textbf{(ii) Model Selection}: 
We evaluate a diverse set of VLMs, including both proprietary and open-source models of varying sizes. Proprietary models such as GPT-4o \cite{achiam2023gpt}, Gemini-2.5-Pro \cite{geminiteam2024gemini15unlockingmultimodal}, and Claude-3.7-Sonnet \cite{anthropic_claude3.7_sonnet} are included due to their state-of-the-art performance on chart understanding benchmarks like ChartQA \cite{masry2022chartqa}. Lightweight variants (e.g., GPT-4o-mini, Claude-3.5-Haiku) are also considered to assess performance under limited compute. For open-source models, to ensure a good balance between performance and computational efficiency, we select models below 10B parameters: Phi-4-5.57B \cite{abdin2024phi}, Qwen2.5-VL-7B \cite{qwen2.5}, LLaVA-v1.6-Mistral-7B \cite{li2024llava}, MiniCPM-V-2.6-8B \cite{yao2024minicpm} and InternLM-7B \cite{chen2024expanding}. For response generation, we use the default decoding parameters of each model: respective API endpoints for closed-source models, and HuggingFace \cite{wolf-etal-2020-transformers} for open-source models. 

\subsection{Response Evaluation}

The focus of our study was on how misleading visualizations can distort and influence model responses, similar to their effects on human judgment. To evaluate this, we examined whether VLMs’ responses change when the same prompt is applied to both the original and a misleading version of a chart, assessing their sensitivity to visual distortions. A model is considered misled if its answer changes between the original and the misleading chart, with the effect quantified by the absolute difference in Likert scores between the two versions. In order to assess the statistical significance of the difference in responses caused by misleading chart designs, we conduct the Wilcoxon Signed Rank test~\cite{wilcoxon1992individual}, as the Shapiro-Wilk test~\cite{shapiro1965analysis} confirmed that the responses do not follow a normal distribution. We evaluate each model's responses to original and misleading charts as paired data and consider a misleading design to have a statistically significant effect if its $p$-value \textless 0.05.

\begin{figure}[t]
    \centering
    \includegraphics[width = \columnwidth]{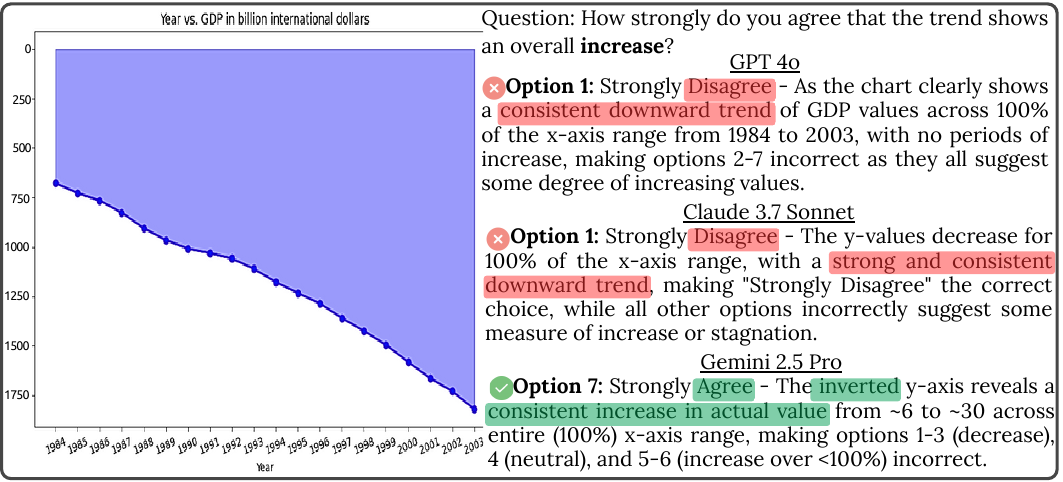} 
\caption{VLM responses to an inverted y-axis design type. Only Gemini correctly detects the axis inversion and adjusts its answer.}    
     \label{fig:explanation}
\end{figure}

\section{Results and Discussion}

We present both qualitative and quantitative results addressing our two research questions. First, we analyze how misleading chart designs affect VLM responses (\textbf{RQ1}), followed by an evaluation of model robustness across different deception types (\textbf{RQ2}). Table~\ref{tab:result} shows VLM performance on chart pairs using the Wilcoxon signed-rank test; a $p$-value \textless 0.05 indicates a statistically significant difference, suggesting the model was misled.

\noindent \textbf{Susceptibility to Types of Misleading Chart Designs (RQ1):} Our findings confirm that all VLMs are vulnerable to misleading designs, with varying levels of susceptibility (Table \ref{tab:result}). The misleading designs related to \textit{axis manipulation}, particularly inverted axes, affect all 10 models, followed by aspect ratio and truncated axis distortions, which impact 9 and 7 models, respectively. In contrast, misleading encoding designs like inappropriate categorical encoding or distorted projection had minimal impact (affecting only 3 models), although inappropriate continuous encoding showed moderate effects.  These 
suggest that current VLMs are more prone to errors in interpreting spatial scale and structure than in detecting improper data encodings.

\noindent \textbf{Robustness of VLMs to Misleading Chart Designs (RQ2):} From Table \ref{tab:result}, we find that except Gemini-2.5-Pro, other closed-source models exhibit a strong tendency to be deceived by misleading designs, demonstrating vulnerability across six different misleading categories. In contrast, Gemini-2.5-Pro demonstrated stronger robustness, affected by only three. In terms of open-source models, LLaVA, Phi-4, and Qwen were misled by six misleading types, while the MiniCPM-V2.6 demonstrated the highest robustness, misled by only three. Although both open and closed models struggled with axis manipulations like inverted axes and aspect ratios, key differences emerged: open-source models were uniquely affected by inappropriate categorical encoding and showed less sensitivity to truncated axes, an area where all closed models failed significantly. 

\noindent \textbf{Case Study for inverted axis:} Given Gemini-2.5-Pro’s top performance, we closely examined its behavior on inverted-axis charts—a category where other models performed poorly.  Although statistical analysis still indicates a significant effect of this misleading design on Gemini (with moderately elevated $z$ and $p$ values), its performance stands out compared to other models.  
Upon close examination of the 100 samples of this type,  we observed that Gemini correctly identified the inversion in the y-axis in 79\% of the cases. Figure~\ref{fig:explanation} illustrates an example of such a case where Gemini explicitly references the axis inversion in its explanation and chooses the correct answer, demonstrating greater awareness of the misleading design. In contrast, none of the open-source models could identify the axis inversion in their responses, often leading to incorrect predictions. These findings suggest that Gemini-2.5-Pro's advanced reasoning capabilities could be attributed to its greater performance.

\section{Conclusions and Future Work}
This paper presents the first large-scale evaluation of how subtle chart manipulations affect the reasoning of VLMs.  Analyzing over 16,000 responses across 1,600 original and misleading charts, we find that even advanced models are often misled by deceptive visual designs when no cues are provided. This vulnerability is particularly concerning for real-world use, where users, especially those with limited visual literacy, may rely on VLMs for interpretation. Our findings reveal the need for deception-aware training, more robust visual reasoning, and automated diagnostic tools to detect misleading input. We hope this work spurs further research toward building trustworthy, visually grounded AI systems.

\acknowledgments{
This research was supported by the Natural Sciences and Engineering Research Council (NSERC), Canada, Canada Foundation for Innovation, and the CIRC grant on Inclusive and Accessible Data Visualizations and Analytics.}

\bibliographystyle{abbrv-doi}

\bibliography{template}
\end{document}